%% file: root.tex
\documentclass[letterpaper, 10 pt, conference]{ieeeconf}  

\IEEEoverridecommandlockouts                              

\usepackage{multirow}
\usepackage{amsmath,amssymb,amsfonts}
\usepackage{xcolor}
\usepackage{siunitx}
\usepackage{booktabs}
\usepackage{subcaption}
\usepackage{flushend}

\newcommand{\ra}[1]{\renewcommand{\arraystretch}{#1}}

\newcommand*\phantomrel[1]{\mathrel{\phantom{#1}}}
\newcolumntype{R}{>{$}r<{$}}

\usepackage{pgfplots}
\pgfplotsset{compat=newest}
\usepgfplotslibrary{groupplots}
\usepgfplotslibrary{dateplot}

\newcommand{\citet}[1]{\citeauthor{#1} \citeyear{#1}}
\usepackage{graphicx}
\usepackage{url}
\usepackage[capitalize]{cleveref}
\usepackage{csquotes}

\usepackage[backend=bibtex,style=ieee,mincitenames=1,maxcitenames=2,maxbibnames=20,url=false,doi=false]{biblatex}
\title{\LARGE \bf
SHAIL: Safety-Aware Hierarchical Adversarial Imitation Learning for Autonomous Driving in Urban Environments$^\dagger$
}

\author{Arec Jamgochian$^{*1}$, Etienne Buehrle$^{*2}$, Johannes Fischer$^2$, and Mykel J. Kochenderfer$^{1}$
\thanks{$^*$Denotes equal contribution}%
\thanks{$^{1}$Stanford University, Stanford, CA 94305 USA
        {\tt\small \{arec, mykel\}@stanford.edu}}%
\thanks{$^{2}$Karlsruhe Institute of Technology, 76131 Karlsruhe, Germany
        {\tt\small \{etienne.buehrle, johannes.fischer\}@kit.edu}}
\thanks{$^\dagger$This material is based upon work supported by the National Science Foundation Graduate Research Fellowship Program under Grant No. DGE-1656518. Any opinions, findings, and conclusions or recommendations expressed in this material are those of the author(s) and do not necessarily reflect the views of the National Science Foundation. 
This work is also supported by the COMET K2---Competence Centers for Excellent Technologies Programme of the Federal Ministry for Transport, Innovation and Technology (bmvit), the Federal Ministry for Digital, Business and Enterprise (bmdw), the Austrian Research Promotion Agency (FFG), the Province of Styria, and the Styrian Business Promotion Agency (SFG). 
}%
}

\begin{document}

\maketitle
\thispagestyle{empty}
\pagestyle{empty}
\begin{abstract}
\input{0-abstract}
\end{abstract}
\input{1-intro}
\input{2-motivation}
\input{3-method}

\input{4-experiments}
\input{5-conclusion}

\input{6-acknowledgment} 




\renewcommand*{\bibfont}{\small}
\printbibliography

\end{document}

%% file: 0-abstract.tex
Designing a safe and human-like decision-making system for an autonomous vehicle is a challenging task. 
Generative imitation learning is one possible approach for automating policy-building by leveraging both real-world and simulated decisions.
Previous work that applies generative imitation learning to autonomous driving policies focuses on learning a low-level controller for simple settings. 
However, to scale to complex settings, many autonomous driving systems combine fixed, safe, optimization-based low-level controllers with high-level decision-making logic that selects the appropriate task and associated controller. 
In this paper, we attempt to bridge this gap in complexity by employing Safety-Aware Hierarchical Adversarial Imitation Learning (SHAIL), a method for learning a high-level policy that selects from a set of low-level controller instances in a way that imitates low-level driving data on-policy.
We introduce an urban roundabout simulator that controls non-ego vehicles using real data from the Interaction dataset. 
We then demonstrate empirically that even with simple controller options, our approach can produce better behavior than previous approaches in driver imitation that have difficulty scaling to complex environments. 
Our implementation is available at \url{https://github.com/sisl/InteractionImitation}.

%% file: 1-intro.tex
\section{Introduction} \label{sec:intro}

The development of autonomous vehicles will greatly impact urban traffic.
Of particular importance is the safety and predictability of autonomous vehicles when interacting with complex environments.
Achieving safe and human-like behavior will require 
a) multiple levels of safety redundancy, 
b) large amounts of real, \enquote{expert} driving data, and 
c) advanced simulators to test behavior before deploying.

Recent reinforcement learning approaches add levels of redundancy to policies learned in simulation by allowing for a hierarchy of control that passes between high-level action selectors and safe low-level optimization-based driving controllers~\cite{mirchevska2021amortized, kamranhierarchical}. 
Though the addition of hierarchical safety layers is intuitive and adds levels of redundancy, the success of any reinforcement learning-based approach hinges on the design of the reward function. A misspecified reward function can be catastrophic.

To resolve the issue of reward misspecification,
imitation learning approaches instead rely on demonstrations from an expert in the environment. 
Data availability invites the use of imitation learning methods that do not interact with the environment (i.e. off-policy methods, such as behavior cloning)~\cite{pomerleau}.
However, these methods suffer from cascading errors when vehicles encounter out-of-distribution states~\cite{ross2011reduction}. 
Some on-policy approaches will query an expert to help guide the learning process safely \cite{ross2011reduction}, 
but querying an expert can be costly or impractical. 
Adversarial imitation learning approaches have been applied with simulators on-policy to circumvent the need for a queryable driving expert \cite{kuefler2017imitating,bhattacharyya2019simulating}, 
but these approaches have mostly been tested in simple driving settings and are still not collision-free.

We approach the safety and environment simplicity limitations of these prior applications of adversarial imitation learning to autonomous driving by taking a hierarchical approach. We note that many autonomous driving systems combine fast, safe, optimization-based controllers for low-level control with high-level logic to select appropriate tasks, controllers, and controller parameters. High-level logic might choose between different options (e.g. \texttt{LaneChangeLeft}, \texttt{Accelerate}, \texttt{TurnRight}, \texttt{EasyBrake}, \texttt{HardBrake}), then pass control to an instance of a low-level controller with the appropriate task and parameters for the chosen option. However, labels for these high-level choices are typically inaccessible in expert trajectories, making direct learning difficult. 

\begin{figure}
    \centering
    \includegraphics[width=0.8\columnwidth]{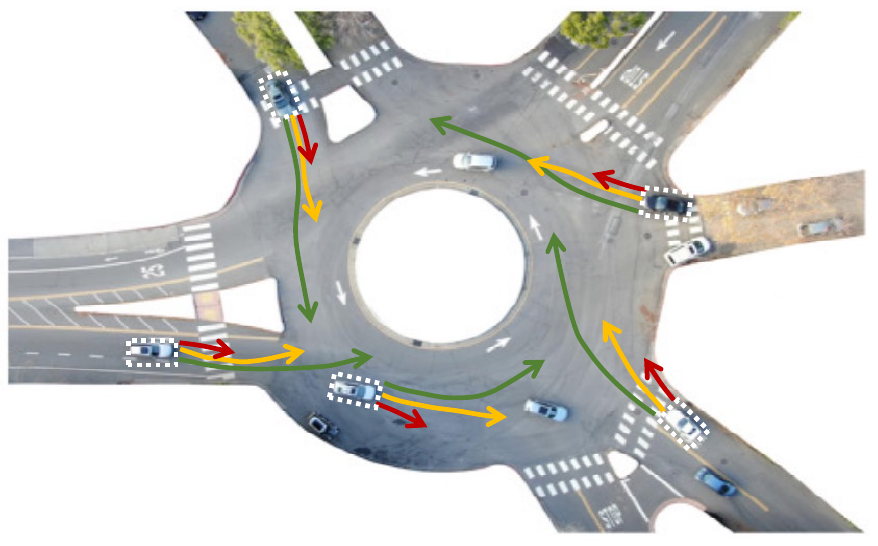}
    \caption{With SHAIL, the ego vehicle learns to choose from a set of safe high-level options to navigate a complex driving environment derived from the Interaction dataset~\cite{interactiondataset}. The learner requires only low-level expert states and actions as opposed to high-level actions or a reward function.}
    \label{fig:set}
\end{figure}

A large body of work exists around hierarchical imitation learning formulations for different robotics problems~\cite{krishnan2016hirl,le2018hierarchical,sharma2018directed,sharma2019third,fox2019multi,jing21a}. In this paper, we employ a method for learning a high-level controller-selection policy that imitates low-level driving data on-policy given a set of \textit{known} low-level controller instances, as is appropriate for autonomous driving. A depiction of our problem setting is shown in~\Cref{fig:set}. We introduce Safety-Aware Hierarchical Adversarial Imitation Learning (SHAIL), which maintains the same low-level occupancy measure-matching objective of previous adversarial imitation learning approaches applied to driving \cite{kuefler2017imitating,ghasemipour2020divergence}, but assumes that low-level data is generated within the options framework \cite{sutton1999between} and reformulates the objective accordingly. Additionally, SHAIL implements a safety awareness layer to adjust the high-level controller selection policy based on active reasoning about the safety or feasibility of different options. 

To demonstrate the effectiveness of SHAIL, we develop a simulator based on real driving data from complex urban driving scenarios in the Interaction dataset \cite{interactiondataset}. Our simulator allows us to adjust the acceleration of an ego vehicle along its real path while other agents in a scene behave according to data. We test a basic implementation of SHAIL in roundabouts, a dynamic driving scenario that is typically difficult for an autonomous vehicle to safely navigate. We compare SHAIL against an IDM adaptation, behavior cloning (BC), non-hierarchical generative adversarial imitation learning (GAIL), and an ablation of SHAIL without the safety layer (HAIL), observing that SHAIL indeed yields safer and more realistic driving behavior.

In summary, the main contributions of this paper are to:
\begin{itemize}
    \item Introduce SHAIL, a methodology for learning a safe high-level action-selection policy that imitates low-level observations and actions, 
    \item Introduce a simulator for complex driving scenarios based on real data, and,
    \item Empirically demonstrate the efficacy of SHAIL compared to IDM, behavior cloning, and non-hierarchical imitation learning.
\end{itemize}

The remainder of this paper is organized as follows: \cref{sec:back} outlines background and related work, \cref{sec:method} describes the methodology for hierarchical adversarial imitation learning with safety constraints, \cref{sec:experiments} presents experiments and results, and \cref{sec:discussion} concludes our research.

%% file: 2-motivation.tex
\section{Background} \label{sec:back}
This section provides necessary background on reinforcement learning, imitation learning, and  hierarchical planning.

\subsection{Reinforcement and Imitation Learning}

Optimal decision-making is often framed in the context of Markov Decision Processes (MDPs). An MDP can be defined as a tuple $\langle \mathcal{S}, \mathcal{A}, T, R, b_0, \gamma \rangle$ and includes a finite state space $\mathcal{S}$, the action space $\mathcal{A}$, a stochastic transition function $T:\mathcal{S} \times \mathcal{A} \times \mathcal{S} \to [0,1]$, a reward function $R: \mathcal{S} \times \mathcal{A} \to \mathbb{R}$, an initial state distribution $b_0: \mathcal{S} \to [0,1]$, and a positive discount factor $\gamma<1$. An MDP policy maps states to a distribution over actions to take $\pi:\mathcal{S} \times \mathcal{A} \to [0,1]$. An optimal policy maximizes expected cumulative discounted reward, $\pi^* \in \arg \max_{\pi \in \Pi}\mathbb{E}_\pi[\sum_{t=0}^\infty \gamma^t R(s_t,a_t) \mid s_0\sim b_0(\cdot)]$. 

In the reinforcement learning setting, the exact transition and reward functions $T$ and $R$ are unknown, but we can interact with an environment to receive generated samples of next state and reward $s', r \sim G(s,a)$. 
There is a body of work in which reinforcement learning is used to generate policies for different autonomous driving scenarios \cite{belletti2017expert,kamranriskaware,chen2019model}. 
These works requires the use of a driving simulator to produce realistic transitions, as well as the manual specification of a reward function. Designing a reward function to capture all desired behavior is extremely difficult, and it is common for learning agents to exploit misspecified reward functions~\cite{amodei2016concrete}. 

In the imitation learning setting, instead of receiving a reward signal, we rely on data in the form of trajectory rollouts from an expert who interacts with the environment. The imitation learning problem can be viewed as a problem of moment-matching between the expert and learner distributions, and methods can broadly be characterized as seeking to match Q-value moments off-policy, Q-value moments on-policy, or reward moments on-policy \cite{swamy2021moments}. In off-policy Q-value moment matching, the learned imitating policy cannot interact with the environment until execution time \cite{Kostrikov2020Imitation}. The most straightforward approach to learn a policy in this setting is through behavior cloning (BC), in which a supervised learner regresses states to actions. This approach has a long history in autonomous driving systems \cite{pomerleau,hawke2020urban}. 

Behavior cloning suffers from an accumulation of errors during testing as an agent ends up in states it has not seen during training, a phenomenon often referred to as covariate shift~\cite{ross2011reduction}. In the on-policy Q-value-matching setting, this accumulation of errors is reduced by introducing a query-able expert who can correct deviating trajectories during training \cite{ross2011reduction}. However, training with a query-able expert can be costly, timely, and impractical. In contrast, on-policy reward moment-matching algorithms assume no access to a query-able expert, but still assume interactions with the environment during training (e.g. using a simulator). 

The state-action occupancy measure under a policy $\pi$ is the (unnormalized) $\gamma$-discounted stationary distribution of states and actions visited under that policy, $\rho^\pi(s,a)= \pi(a\mid s)\rho^\pi(s)$, where $\rho^\pi(s)=\sum_{t=0}^\infty \gamma^t P(s_t=s \mid \pi)$. We can similarly define the state-action occupancy measure of the expert policy, $\rho^\text{exp}(s,a)$. One perspective formulates imitation learning as moment-matching between expert and learned occupancy measures, done by minimizing some $f$-divergence between the associated distributions $\min_\pi D_f((1-\gamma)\rho^\text{exp}(\cdot,\cdot) \|(1-\gamma) \rho^\pi(\cdot,\cdot))$ \cite{ghasemipour2020divergence}. 
In the on-policy reward matching setting, this objective can be written as a two-player game between a policy generator $\pi_\theta$ and an observation-action discriminator $D_\phi$:
\begin{multline} \label{eq:fgail}
    \min_{\pi_\theta} \max_{D_\phi} \mathbb{E}_{(s,a)\sim\rho^\text{exp}(\cdot,\cdot)} [D_\phi (s,a)] \\ - \mathbb{E}_{(s,a)\sim\rho^{\pi_\theta}(\cdot,\cdot)} [f^* (D_\phi (s,a))]\text{,}
\end{multline}
where $f^*$ denotes the convex conjugate of $f$. This objective can be optimized by alternating between discriminator gradient ascent steps to optimize the discriminator parameters $\phi$ and policy gradient ascent steps to optimize the parameters $\theta$ of a stochastic policy. This latter step can be viewed as reinforcement learning with an `imagined' reward signal of $r(s,a) = f^* (D_\phi (s,a))$. These steps use Monte Carlo methods (and a replay buffer) to estimate the expectations \cite{ho2016generative,fu2018learning}. 

These generative methods have been used to imitate highway driving behavior \cite{kuefler2017imitating}. Later work improves upon this by augmenting the learned reward model with soft constraints to avoid bad states and actions \cite{bhattacharyya2019simulating}. Two shortcomings of these works are that they 
a) mostly consider highway driving, a relatively simple driving scenario, and
b) only learn low-level controllers, for which safety is more difficult to guarantee in comparison to traditional optimization-based controllers.

\subsection{Hierarchical Planning}

Human driving in complex environments is naturally hierarchical. One can model hierarchical planning through the context of options \cite{sutton1999between}, in which a low-level controller $o$ is chosen from a finite set of options $\mathcal{O}$ and executed until termination, upon which a new valid option is chosen. An options model can be defined with the tuple $\langle\mathcal{S}, \mathcal{A}, \{\mathcal{I}_o, \pi^L_o, \beta_o\}_{o \in \mathcal{O}}, T, R, b_o, \gamma \rangle$. In addition to the components of an MDP, an options model defines $K$ options (indexed by $o$) which each define a set of states from which they can be initialized $\mathcal{I} \subseteq \mathcal{S}$, a low-level control policy $\pi^L: \mathcal{S} \times \mathcal{A}\to [0,1]$, and the probability that the option will be terminated from any given next state $\beta: \mathcal{S} \to [0,1]$. A high-level policy over options denotes the probability of choosing from the valid set of options in any given state $\pi^H(o \mid s)$, where $s \in \mathcal{S}$ and $o \in \mathcal{O}$ such that $s \in \mathcal{I}_o$.

Recent work considers hierarchical reinforcement learning for planning in driving scenarios \cite{chen2018deep,kamranhierarchical,mirchevska2021amortized}. While still suffering from the pitfalls of a manually specified reward function, these approaches have the benefit that a high-level action-selector can hand over control to safe, low-level, optimization-based planners.
\citeauthor{mirchevska2021amortized} use this approach to learn a high-level controller that can choose safe gaps in highway traffic for an optimization-based low-level controller to navigate to~\cite{mirchevska2021amortized}. In their approach, the high-level controller only targets reachable gaps, while if a targeted gap no longer is reachable during low-level execution, control is passed back to the high-level controller. 

Additional work considers hierarchical approaches to imitation learning. For example, \citeauthor{henderson2018optiongan} perform imitation learning hierarchically (as opposed to hierarchical imitation learning) by learning multiple generators and discriminators to match low-level data, and learning a mixture-of-experts policy over those generators to follow~\cite{henderson2018optiongan}. \citeauthor{le2018hierarchical} describes a formulation to perform hierarchically-guided behavior cloning and dataset aggregation, however, this assumes labeling of the high-level option, which we do not~\cite{le2018hierarchical}. 

\citeauthor{jing21a} perform hierarchical on-policy reward moment-matching by framing an objective to match the moments over states, actions, and options and alternating between expert option label inference and joint policy training. In this work, we keep low-level options fixed, as is more appropriate for driving. 
As a result, our work is in line with moment-matching objectives over state and action~\cite{ghasemipour2020divergence} as opposed to those additionally over options~\cite{jing21a}. This can be viewed as a subclass of~\citeauthor{jing21a} \cite{jing21a} where there is no need to infer latent options in the data, or as an extension of~\citeauthor{ghasemipour2020divergence} \cite{ghasemipour2020divergence} in which the state-action pairs are drawn hierarchically. 

Much of recent work that performs vehicle behavior prediction hierarchically (e.g. ~\cite{wang2022transferable}) can be easily extended to off-policy hierarchical imitation learning, as many policies learned to predict vehicle behavior when conditioned on a goal could be extended to control an ego vehicle. 
A flavor of on-policy hierarchical imitation learning has been applied to driving policies, in which long-horizon planning learned to mimic a query-able expert is interleaved with fast, short-horizon, low-level optimal control~\cite{sun}. In contrast, we propose learning a high-level controller on-policy that imitates low-level data without a query-able expert, and characterize a broader class of high-level control options. 

%% file: 3-method.tex
\section{Methodologies} \label{sec:method}

This section formulates SHAIL, first by reformulating the occupancy measure-matching objective for a policy that generates low-level data hierarchically, and then by designing a safety-aware high-level controller. 
\subsection{Hierarchical Adversarial Imitation Learning} \label{sec:hail}

We first formulate the occupancy measure-matching objective in~\Cref{eq:fgail} for a policy that is generating states and actions hierarchically. We do this by expanding the occupancy measure over options that would lead to state $s$ and action $a$ during their execution, and states in which the options begin executing. We expand over the initiation states $s_\tau=h$ that begin executing the options $o$ at time $\tau$ under which low-level states and actions $s$ and $a$ can be observed at time $t$: 
\allowdisplaybreaks
\begin{align}
&\phantomrel{=}{} \rho^{\pi}(s,a) = \sum_{t=0}^\infty \gamma^t P(s_t=s, a_t=a)   \\
&= \sum_{h,o}\sum_{t=0}^\infty\sum_{\tau=0}^t  \gamma^\tau P(s_\tau=h, o_\tau=o) \nonumber \\ 
& \phantomrel{======}{}\cdot\gamma^{t-\tau} P(s_t=s, a_t=a \mid s_\tau=h, o_\tau=o) \\
&= \sum_{h,o}\sum_{\tau=0}^\infty\gamma^\tau P(s_\tau=h, o_\tau=o) \nonumber \\ 
& \phantomrel{====}{}\cdot\sum_{t=\tau}^\infty \gamma^{t-\tau} P(s_t=s, a_t=a \mid s_\tau=h, o_\tau=o) \\
&= \sum_{h,o}\sum_{\tau=0}^\infty\gamma^\tau P(s_\tau=h, o_\tau=o) \nonumber \\ 
& \phantomrel{====}{}\cdot\sum_{t=0}^\infty \gamma^{t} P(s_t=s, a_t=a \mid s_0=h, o_0=o) \\
&= \sum_{h,o} \rho^{\pi^H}(h,o) \rho^{\pi^L}(s, a \mid h, o)\text{,}
\end{align}
where $\rho^{\pi^H}(h,o) = \pi^H(o \mid h)\sum_{\tau=0}^\infty \gamma^\tau P(s_\tau=h)$ 
is the discounted occupancy measure for ending up in a high-level state $h$ and initiating option $o$, and $\rho^{\pi^L}(s, a \mid h, o) = \pi_o^L(a \mid s)\sum_{t=0}^\infty \gamma^{t} P(s_t=s\mid s_0=h, o_0=o)$ 
is the discounted occupancy measure of states and actions under an option $o$ which was initialized in state $h$ at time $0$.

We apply this hierarchical representation of occupancy measure $\rho^\pi(s,a)$ to reformulate the measure-matching objective in~\Cref{eq:fgail} for policy data that is generated hierarchically:
\begin{align}
    & \min_{\pi} \max_{D_\phi} \mathbb{E}_{(s,a)\sim\rho^\text{exp}(\cdot,\cdot)} [D_\phi (s,a)] - \sum_{s,a} \rho^{\pi}(s,a) r(s,a) \\
    & \min_{\pi^H_{\theta}} \max_{D_\phi} \mathbb{E}_{(s,a)\sim\rho^\text{exp}(\cdot,\cdot)} [D_\phi (s,a)] \nonumber \\ 
    & \phantomrel{=====} {} - \sum_{h, o} \rho^{\pi^H_{\theta}}(h, o) \sum_{s, a} \rho^{\pi^L}(s, a \mid h, o)  r(s,a) \\
    & \min_{\pi^H_{\theta}} \max_{D_\phi} \mathbb{E}_{(s,a)\sim\rho^\text{exp}(\cdot,\cdot)} [D_\phi (s,a)] - \mathbb{E}_{(h,o)\sim\rho^{\pi^H_{\theta}}(\cdot,\cdot)} [\tilde{r} (h,o)]\text{,}
\end{align}
where $\tilde{r} (h,o) = \mathbb{E}_{(s,a)\sim\rho^{\pi^L}(\cdot, \cdot \mid h, o)}  [f^* (D_\phi (s,a))]$. 

Optimizing this objective can be done in a fashion similar to optimizing the objective in~\cref{eq:fgail}. Discriminator updates remain identical, while generator updates require performing policy gradients on $\pi_\theta^H(o \mid h)$ where the new \enquote*{imagined} high-level reward $\tilde{r} (h,o)$ accumulates the discounted low-level \enquote*{imagined} discrimination rewards from the execution of the chosen option. That is, $\tilde{r} (h,o)$ can be estimated as 
$\sum_{t=0}^{T_o} \gamma^t r(s_t, a_t)$, where $s_0 = h, a_t \sim \pi^L_o(\cdot|s_t)$, and $T_o$ is the duration of the option.

\subsection{Safety-Aware Hierarchical Adversarial Imitation Learning} \label{sec:safety}

Many practical implementations of policy gradients rely on a fixed-size action space \cite{schulman2015trust,schulman2017proximal}. Because of this restriction, we are limited to an option set where any option can be initialized from every state, i.e. $\mathcal{I}_o=\mathcal{S}$ for all $o \in \mathcal{O}$. This assumption can be very limiting in terms of safety. Often times, we have information about restricted options from different states (e.g. an \texttt{Accelerate} option should not be taken from a red light). Additionally, we might be able to make predictions about the safety of different controllers. For example, this can be done strictly with formulations of reachability of a controller, or more loosely through notions of scene understanding (e.g. \enquote*{it is probably unsafe to make a turn since there are vehicles crossing the intersection}). 
SHAIL improves upon the formulation of hierarchical adversarial imitation learning presented in the previous section by designing a high-level option-selection policy that incorporates sensitivity to option safety. 

Safety awareness is incorporated by assuming that the agent can reason about the safety or availability of different options from different states. We introduce a binary random variable $z$ which predicts safety or availability of a low-level controller, denoting the probability that an option $o$ is safe when executed from a high-level state $s$ as $p_\text{safe}(z^1\mid s,o)$. This allows us to design the options such that control is passed back to the high-level option selector according to this safety prediction, i.e. $\beta_o(s) = 1 - p_\text{safe}(z^1\mid s,o)$. 
This option termination formulation expands on the one used by~\citeauthor{mirchevska2021amortized} in the hierarchical reinforcement learning setting~\cite{mirchevska2021amortized} to admit probabilistic controller safety predictions rather than binary controller availability.

With this formulation, we additionally design a high-level controller that conditions on controller safety:
\begin{equation}
\pi^H_\theta(o\mid s, z^1) 
\propto p(o, z^1 \mid s) 
=  p_\text{safe}(z^1 \mid s,o) \psi^H_\theta(o \mid s)\text{,}
\end{equation}
where $\psi^H_\theta$ is a learnable controller selector. This high-level controller reweights options based on predictions of their safety or availability. 
It can be easily shown that learning with this substituted policy is equivalent to minimizing the divergence to an agent occupancy measure conditioned on safety, $\rho^\pi(s,a\mid z^1)$. 
This scheme requires at least one option with nonzero safety probability (e.g. a permanent \enquote*{safe} controller), otherwise the high-level policy will not represent a valid distribution over controllers. Additionally, to learn a useful option selector, the options should have some semantic meaning that holds across different initialization states.

Learning $\psi^H_\theta$ with policy gradients on this policy formulation requires storing the safety probabilities seen during option initiation in the replay buffer. That is, for each option $o$ initiated from a state $h$, the replay buffer consists of samples of the form
$(h, o, p_\text{safe}(z^1 \mid h, \cdot), \tilde{r}(h,o))$.

%% file: 4-experiments.tex
\section{Experiments} \label{sec:experiments}

Our experiments demonstrate the effective use of SHAIL in a driving simulator. We introduce our own simulator based on real data in urban driving environments, and demonstrate the improvements regarding safety that can be achieved in comparison with baseline models, even by a simple SHAIL implementation.

\subsection{Setup} \label{sec:setup}

\subsubsection{Simulator}

To test this approach in a more complicated driving environment, we introduce the Interaction simulator.\footnote{\url{https://github.com/sisl/InteractionSimulator}} The Interaction simulator is an OpenAI Gym \cite{brockman2016openai} simulator that uses underlying data from the Interaction dataset of complex urban driving scenes \cite{interactiondataset}. 
The dataset consists of recorded track files from driving scenarios in different urban driving situations like roundabouts or intersections.

The simulator itself fixes vehicle paths and spawn times based on the recorded data in the Interaction dataset, and admits control of vehicle accelerations along the path. This is a reasonable navigation strategy, as it may be common for a separate module to determine the path of a vehicle navigating a complex scene. 

In our experiments\footnote{\url{https://github.com/sisl/InteractionImitation}}, we focus on controlling a single vehicle that is modeled with double integrator dynamics and moves along its recorded path while non-ego vehicles follow their recorded trajectories. Simulations are terminated when the vehicle leaves the scene. 

\subsubsection{Features}

We assume that the ego vehicle encodes its absolute velocity, yaw rate, and lidar-like measurements of the relative position and velocity of the closest vehicle in each $72^{\circ}$ sector of its surroundings. We use this very simple subset of autonomous vehicle features in order to avoid overfitting to our small dataset. 

\subsubsection{Models}
We evaluate the following baseline models in our experiments:

\textbf{Expert}: The expert model uses the default accelerations from the Interaction dataset.

\textbf{IDM}: The Intelligent Driver Model (IDM) models vehicle accelerations in fixed-lane driving when following a vehicle~\cite{treiber2000congested}. This model is problematic for uncontrolled driving, where it is not clear which vehicle the ego should `follow'. 
We choose the follow vehicle as the closest vehicle that lies within two meters of the ego's planned path and has less than a $30^{\circ}$ difference in heading. 
We target a desired speed of 8.94 m/s (20 mph), a minimum spacing of 3 m, a desired time headway of 0.5 s, a nominal acceleration of 3 m/s$^2$, and a comfortable braking deceleration of 2.5 m/s$^2$.

\textbf{BC}: We implement a behavior cloning agent that directly regresses features to a mean and standard deviation parameterizing a normal distribution for ego vehicle acceleration. Our model is a feedforward neural network with layers of fixed hidden size. We train our model by minimizing the negative log-likelihood of expert actions under the action distribution predicted from their preceding states. 

\textbf{GAIL}: We compare against a model learned through Generative Adversarial Imitation Learning (GAIL)~\cite{ho2016generative,kuefler2017imitating}. Both discriminator and policy models are feedforward neural networks, the latter again outputting parameters for a normal distribution over next action. We use the same optimization objective as~\citeauthor{ho2016generative}, as well as proximal policy optimization (PPO)~\cite{schulman2015trust} to learn a policy. We do not compare recurrent policies, as our episodes are too short to compare against recurrent hierarchical policies.

\textbf{SHAIL}: To demonstrate the effectiveness of SHAIL, we formulate a \textit{very simple} hierarchical policy. Our high level controller chooses from a set of options which target a particular velocity at a particular future time, $\mathcal{O} = \{(v,t) \mid v \in \mathcal{V},\ t \in \mathcal{T}\}$, where $\mathcal{V}$ and $\mathcal{T}$ are discrete velocity and time sets. 
The low-level controller for each option commands a fixed acceleration to bring the vehicle to the desired velocity at the desired time. The safety predictor returns a binary indicator for whether the option is scheduled to collide with other vehicles if they maintain their velocity. 
Additionally, we overwrite the largest deceleration option to always be valid, thereby rendering it a default \enquote*{safe} option \texttt{HardBrake}. Again, we use the objective from~\citeauthor{ho2016generative}, and PPO for policy gradients. We also learn a version of SHAIL without the safety layer or early option termination for ablation (\textbf{HAIL}).

\subsubsection{Training and Metrics}

Our experiments focus on model performance in roundabouts, a customarily tricky scenario for an autonomous system to navigate. Specifically, we look at the \texttt{DR\_USA\_Roundabout\_FT} scene, which includes five recordings consisting of over 750 ego vehicle tracks. We believe it to be the most difficult roundabout scene in the dataset. Upon collision of a controlled vehicle, the environment is reset with a new random vehicle. 

We perform two experiments. In our first experiment (\textit{in-distribution}), we train and test models in the same environment, which selects vehicles only from the first track file. The purpose of this experiment is to compare absolute potential model performance. This in-distribution testing corresponds with what was done by~\citeauthor{kuefler2017imitating}~\cite{kuefler2017imitating}.

In our second experiment (\textit{out-of-distribution}), we train and validate in an environment that randomly selects vehicles from scene recordings 1--4, and we report metrics on scene 5. This out-of-distribution testing evaluates how the models perform on unseen vehicle data, though we acknowledge that we are still operating in the same driving setting. In both experiments, hyperparameters (e.g. model architecture, options sets, etc.) are optimized by choosing the ones which yield the highest success rate in the training environment. Please refer to our code for all parameters.

To avoid collisions caused by non-ego vehicles following their recorded trajectories, our test environment overrides non-ego accelerations with the IDM policy in case of an impending not-at-fault collision. For each model, we report success rate (the rate at which an episode does not terminate in collision), the average travelled distance (m), the root mean square error in position to the expert measured at 10 seconds into the trajectory (m), the average absolute difference between average speed of each vehicle under expert and modelled control (m/s), and the Jensen-Shannon divergence between the distributions over all accelerations. The divergence is estimated by fitting a histogram distribution to the two sets of samples.  

\subsection{Results and Discussion} \label{sec:results}

\begin{figure}[h]
    \centering
    \includegraphics[width=0.8\columnwidth]{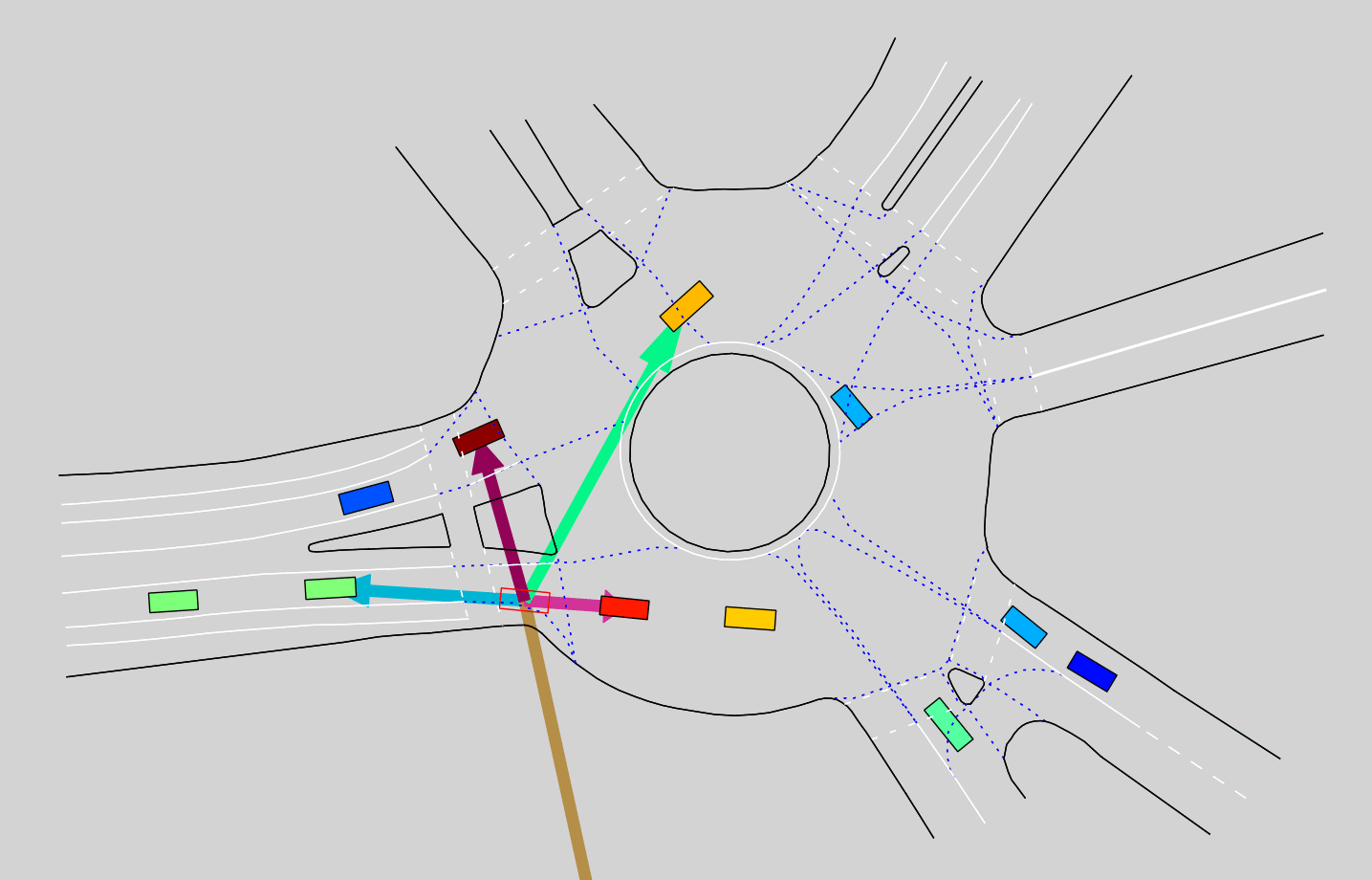}
    \caption{A single time-step of a policy learned by SHAIL interacting with the environment. The ego vehicle has available to it its own motion state and lidar-like measurements capturing relative state information about up to five surrounding vehicles.}
    \label{fig:example}
\end{figure}

\begin{table}[htpb]

\ra{1.2}
\caption{A comparison of performance between expert, IDM, BC, GAIL, HAIL (ablation), and SHAIL policies in both in-distribution (above) and out-of-distribution (below) roundabout experiments. We report metric means and two standard deviations after training each model five times.}
\centering
\scalebox{0.84}{\input{figs/datatable-idm}}
\label{table:results}
\end{table}

Our simulation environment is visualized in~\Cref{fig:example}, while results from multiple runs of both in- and out-of-distribution roundabout experiments are shown in~\Cref{table:results}. 
From the success rate and average distance traveled metrics, we can see that the simple IDM rule-following policy, behavior cloning, and GAIL all have trouble successfully navigating through the roundabout. 
From the in-distribution experiment, we see that incorporating safe options, even those as simple as the ones we suggest, can yield better performance when driving in complex environments. 
We see improvements in the success rate and travel distance metrics. We see that metrics that judge similarity to expert position and speed perform comparably well in both GAIL and SHAIL, indicating some human-like behavior in both.

We note that the distribution over SHAIL accelerations is quite far from the expert distribution, especially when compared to GAIL. This disparity can be attributed to our overly simple option design. Our controllers attempt to target different velocities at different points by holding fixed accelerations, ultimately resulting in jerky behavior. We could bridge this gap by implementing more comfortable, human-like controller options.  

In our ablation study, implementing the same options without any safety layering (HAIL) results in a severe drop in performance. Intuitively, when implementing options without safety or termination criteria, we are reducing the space of immediate actions available to the agent. Even if we could learn a good imitating controller that predicts which options might be safe from a particular state, we have no method for terminating if the options become unsafe. In our experiments, this is made even worse by our simple controllers, which stick to the action plan that was initiated during option selection and do not adjust their plan based on environment feedback. 

We see similar results in our out-of-distribution experiment, noting that the performance gap between SHAIL and other models is even greater. Though none of the learning methods perform as well as they would in-distribution, this performance gap suggests that SHAIL could be a good approach for navigating new situations. We note though that our out-of-distribution experiment tested in the same roundabout setting as training, just with new vehicle data. Though we believe this to be the hardest setting, it would be interesting to train our approach over different settings and test on a fully unseen one to see how well the learned safe option selector could generalize.

%% file: figs/datatable-idm.tex
\begin{tabular}{@{}r RRRRR @{}}
\toprule
\textbf{Model} & {\textbf{Success \%}} &{\textbf{Travel Dist.}} & {\textbf{RMSE$_{10s}$}} & {\textbf{$|\Delta V_\text{avg}|$}}  & {\textbf{{Accel. JSD}}}\\
\midrule
\texttt{Expert}  & 100 & 82.1 & $---$  & $---$ & $---$ \\
\texttt{IDM}         & 66.2& 67.5 & 19.2 & 1.52& 0.050 \\
\texttt{BC}         & 45.3 \scriptstyle\pm 3.0 & 49.8 \scriptstyle\pm 1.4 & 22.0 \scriptstyle\pm 1.9 & 1.88 \scriptstyle\pm 0.11 & 0.275 \scriptstyle\pm 0.017 \\
\texttt{GAIL}       & 68.3 \scriptstyle\pm 4.8 & 65.3 \scriptstyle\pm 3.9 & 14.9 \scriptstyle\pm 1.1 & \mathbf{1.08} \scriptstyle\pm 0.14 & \mathbf{0.016} \scriptstyle\pm 0.021 \\
\texttt{HAIL} (ab) & 53.0 \scriptstyle\pm 24.1 & 54.5 \scriptstyle\pm 15.3 & 18.3 \scriptstyle\pm 4.8 & 1.87 \scriptstyle\pm 0.86 & 0.333 \scriptstyle\pm 0.008 \\
\texttt{SHAIL}       & \mathbf{77.7} \scriptstyle\pm 3.1 & \mathbf{70.6} \scriptstyle\pm 2.3 & \mathbf{14.7} \scriptstyle\pm 2.1 & 1.27 \scriptstyle\pm 0.23 & 0.312 \scriptstyle\pm 0.012 \\
\midrule
\texttt{Expert}      & 100 & 81.9 & $---$  & $---$ & $---$\\
\texttt{IDM}         & 56.3& 59.9 & 21.0 & 1.45& 0.061 \\
\texttt{BC}         & 40.4 \scriptstyle\pm 3.1 & 48.8 \scriptstyle\pm 1.3 & 22.5 \scriptstyle\pm 0.5 & 1.92 \scriptstyle\pm 0.05 & 0.290 \scriptstyle\pm 0.018 \\
\texttt{GAIL}       & 51.6 \scriptstyle\pm 6.8 & 54.5 \scriptstyle\pm 4.7 & \mathbf{14.7} \scriptstyle\pm 1.6 & 1.41 \scriptstyle\pm 0.21 & \mathbf{0.031} \scriptstyle\pm 0.018 \\
\texttt{HAIL} (ab)      &  39.9 \scriptstyle\pm 8.6 & 46.3 \scriptstyle\pm 6.5 & 21.1 \scriptstyle\pm 5.0 & 2.10 \scriptstyle\pm 0.55 & 0.328 \scriptstyle\pm 0.016 \\
\texttt{SHAIL}       & \mathbf{64.4} \scriptstyle\pm 6.8 & \mathbf{61.6} \scriptstyle\pm 3.6 & 18.1 \scriptstyle\pm 1.8 & \mathbf{1.37} \scriptstyle\pm 0.34 & 0.300 \scriptstyle\pm 0.027 \\
\bottomrule
\end{tabular}

%% file: 5-conclusion.tex
\section{Conclusion} \label{sec:discussion}

Previous work applying adversarial imitation learning to autonomous driving focuses on learning low-level control policies. However, since many autonomous driving systems rely on optimization-based control to provide safe low-level policies, it may be more prudent to rely on data-driven tools to learn high-level control policies. 
In this work, we introduced Safety-Aware Hierarchical Adversarial Imitation Learning (SHAIL), a methodology for learning high-level control policies in a simulator such that low-level expert trajectories are imitated. SHAIL incorporates an additional layer to reason over option safety and availability, allowing to guarantee that only safe and feasible actions are executed. To demonstrate our approach, we developed a simulator based on the Interaction dataset of real complex urban driving scenarios. Finally, we compared SHAIL to previous approaches to empirically demonstrate the safety improvements that it affords when navigating a roundabout.

SHAIL inherits all of the limitations from generative adversarial networks, policy optimization, and simulation-based approaches to policy learning. Additionally, SHAIL limits high-level options to a fixed set of predetermined low-level control policy instances. These options must hold semantic meaning in different initialization states for meaningful learning to occur. One limitation of our experiments is the simplicity of the low-level controllers used, which are meant only to demonstrate the potential of our proposed method. More advanced low-level controllers, safety predictors, and termination criteria can easily be substituted into our method. SHAIL can be extended to perform reward augmentation to design policies that avoid known unfavorable behavior. It can also potentially be applied across settings to learn more generalizable option-selection.

%% file: 6-acknowledgment.tex
\section*{Acknowledgment}

The authors acknowledge Kunal Menda for early work on the Interaction simulator.